\documentclass[conference]{IEEEtran}
\IEEEoverridecommandlockouts

\usepackage{cite}
\usepackage{amsmath,amssymb,amsfonts}
\usepackage{algorithmic}
\usepackage{graphicx}
\usepackage{textcomp}
\usepackage{xcolor}
\usepackage{array}
\usepackage{bm}
\usepackage{multirow}
\def\BibTeX{{\rm B\kern-.05em{\sc i\kern-.025em b}\kern-.08em
    T\kern-.1667em\lower.7ex\hbox{E}\kern-.125emX}}
\begin{document}

\title{PointLCA-Net: Using Point Clouds for Energy Efficient Sparse Spatio-Temporal Signal Recognition in Neuromorphic Systems  \\
}

\author{\IEEEauthorblockN{Sanaz Mahmoodi Takaghaj}
 \IEEEauthorblockA{\textit{Department of Computer Science and Engineering, and} \\
 \textit{School of Engineering Design and Innovation}\\
 \textit{Penn State University}\\
 University Park, PA USA \\
 sxm788@psu.edu}}

\maketitle

\begin{abstract}
In recent years, the field of neuromorphic systems has gained significant traction as the demand for AI applications in energy-constrained autonomous systems, such as unmanned aerial vehicles (UAVs) and robotics, continues to grow. The advent of Dynamic Vision Sensor (DVS) and other event-based sensors, which generate sparse spatio-temporal signals, has propelled the field of neuromorphic engineering. This advancement calls for the development of new machine learning models or the refinement of existing ones to effectively leverage neural dynamics and exploit spike-based communication among neurons. These sparse spatio-temporal signals can be treated as 3D point sets (point clouds), representing the space and time at which these events occur, and can be processed using PointNet architectures. However, applying deep learning techniques such as backpropagation and deploying these models on energy-constrained neuromorphic devices presents challenges in computational resources and memory, as well as real-time processing and power consumption. This work presents PointLCA-Net, which leverages the strengths of PointNets for extracting robust features from input point sets, while utilizing the efficiency of Exemplar LCA to encode these features. PointLCA-Net achieves a high accuracy of 93.41\% on DVS128 while reducing energy consumption by approximately 92\% compared to other spiking neural networks applied to point clouds, making the system more efficient and well-suited for energy-constrained applications.
\end{abstract}


\begin{figure*}
\centering
\includegraphics[scale=0.12]{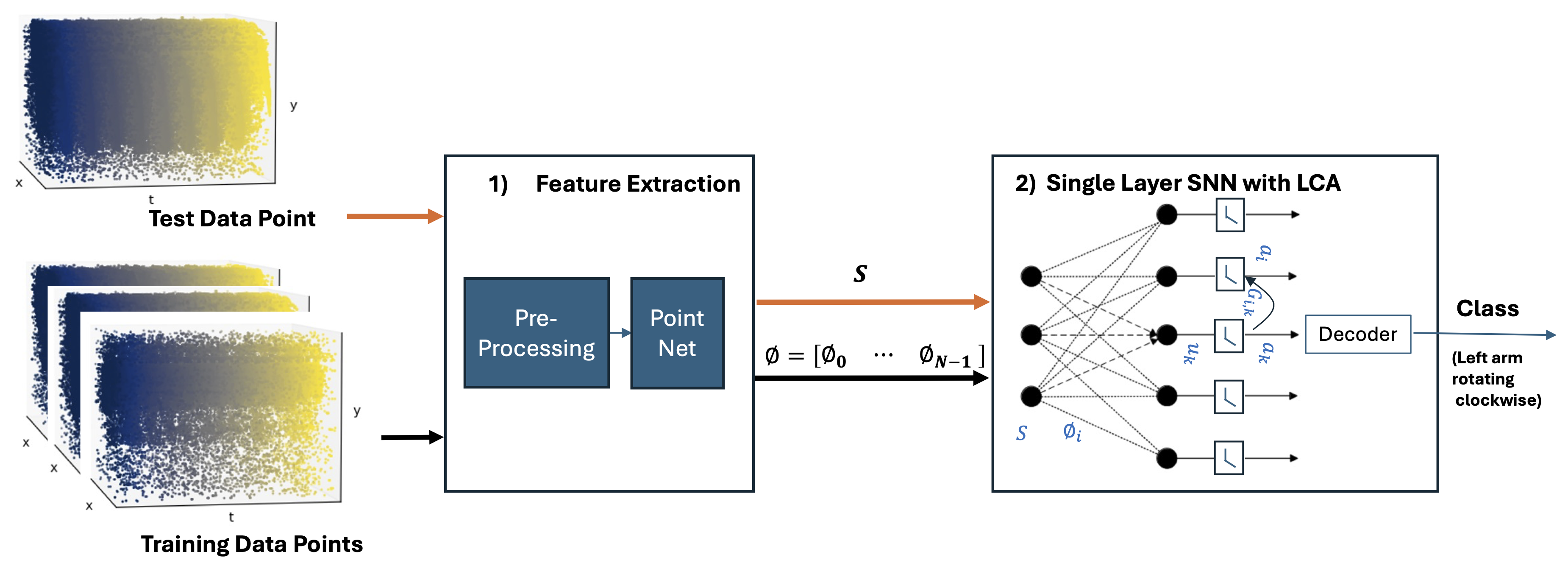}
\caption{PointLCA-Net Architecture: Features ($\phi_i$) are extracted from the training data and stored in the synaptic weights of the Exemplar LCA-Decoder. The orange arrows represent the inference process, which follows the completion of training (feature extraction).}
\label{lca-arch}
\end{figure*}

\section{Introduction}
\label{sec:introduction}
Deep learning has significantly advanced the analysis of complex data structures, with architectures such as PointNet~\cite{qi2017pointnet} and PointNet++~\cite{qi2017pointnet++} revolutionizing the processing of three-dimensional (3D) point sets (point clouds). These architectures have led to substantial improvements in tasks such as 3D object classification and segmentation~\cite{qi2017pointnet, qi2017pointnet++, ma2022rethinking}. PointNet takes point clouds as input, applies input and feature transformations, and aggregates the point features using a max pooling function. The network outputs a set of classification scores for \( k \) classes, along with segmentation scores. PointNet++ is a hierarchical extension of PointNet that applies the architecture recursively to a nested partitioning of the input set and adaptively combines features from multiple scales to enhance the generalizability of PointNet to complex 3D shapes and point cloud structures. Unlike PointNet, which processes points globally, PointNet++ processes points within local neighborhoods, performing additional clustering and multi-scale feature extraction. 

Point clouds can represent discrete spatio-temporal signals by capturing both spatial and temporal information as discrete points in space-time. Wang et al.~\cite{wang2019space} proposed a method for constructing Space-time Event Clouds from Dynamic Vision Sensors (DVSs) and applied the PointNet and PointNet++ architectures to gesture recognition task on DVS128~\cite{amir2017low}. Despite these advancements, applying such techniques on edge devices — where energy efficiency is crucial — remains challenging due to the limitations of conventional computing systems. These challenges include issues related to data movement, memory capacity, and power consumption. Conventional processors, such as CPUs and GPUs/TPUs, are optimized for handling dense, synchronized data structures through efficient instruction and data pipelines. However, their performance significantly deteriorates when dealing with sparse, asynchronous event streams~\cite{dave2021hardware}. This calls for a departure from traditional von Neumann computing systems, with neuromorphic computing serving as one potential solution.

Recently, neuromorphic systems -- drawing inspiration from the human brain's capabilities and/or structures - have emerged as a transformative approach in computing systems, offering energy-efficient and massively parallel processing elements. A new wave of neuromorphic processors including TrueNorth~\cite{merolla2014million, amir2017low}, Loihi~\cite{davies2018loihi, davies2021advancing}, SpiNNaker~\cite{furber2014spinnaker, furber2016large, hoppner2021spinnaker}, BrainScaleS~\cite{schmitt2017neuromorphic,pehle2022brainscales}, Tianjic~\cite{pei2019towards}, and DYNAP-SE~\cite{richter2024dynap} has been developed to deliver low-latency and low-power consumption, making them well-suited for modeling Spiking Neural Networks (SNNs)~\cite{ghosh2009spiking, gruning2014spiking} and processing sparse data streams from event-based sensors. 

SNNs represent a compelling paradigm for modeling temporal dynamics, providing advantages in processing sequential data through discrete spike events that emulate biological neural activity. SNNs have the potential to excel at representing and processing temporal sequences, making them particularly well-suited for tasks involving dynamic patterns and time-varying signals.

Neuromorphic systems can efficiently execute Vector-Matrix Multiplication (VMM) operations, also known as Multiply-and-Accumulate (MAC) operations, by leveraging a crossbar array of memory elements~\cite{aguirre2024hardware}, which are fundamental to deep learning algorithms. Non-volatile memory elements, such as PCM~\cite{kim2015nvm, kuzum2012nanoelectronic, tuma2016detecting} and RRAM~\cite{covi2016analog, pantazi2016all, rao2023thousands, serb2016unsupervised}, have been proposed for integration into synapse and neuron circuits~\cite{sebastian2020memory}. While most efforts to apply in-memory computing to SNNs have focused on unsupervised learning with local learning rules, such as Spike-Timing Dependent Plasticity (STDP)~\cite{indiveri2015memory}, which adjusts synaptic weights based on the relative timing between inputs and outputs, these approaches encounter significant challenges when applied to complex datasets and tasks. STDP-based learning rules have limited scalability and are not well-suited for more complex tasks, such as recognizing spatio-temporal signals. As a result, there is a clear need for new learning rules that can handle dynamic data and address these complex challenges.

A particularly interesting model in neuromorphic computing is the Exemplar LCA-Decoder\cite{mahmoodi2024d, takaghaj2024vit}, which builds upon the Locally Competitive Algorithm (LCA)\cite{4379981, rozell2008sparse}. The Exemplar LCA-Decoder is a computational model and learning algorithm that iteratively updates neuron activity to achieve a sparse representation of input data. As a single-layer encoder-decoder architecture, it first encodes the input data features by identifying a sparse representation through neuron activity and then uses this sparse representation for classification tasks during the decoding phase. A key component of this algorithm is the construction of a comprehensive dictionary of input data features. 

This work builds upon the foundational principles of the Exemplar LCA-Decoder and introduces a novel algorithm, PointLCA-Net, to address the challenges of recognizing spatio-temporal signals on neuromorphic platforms. We propose a two-stage approach in which spatio-temporal data, such as that from DVS cameras, is initially preprocessed and converted into point clouds. In the first stage, a pre-trained PointNet generates robust feature representations from this 3D point cloud data. These features, which capture both geometric and semantic information, are then input into the Exemplar LCA-Decoder architecture in the second stage for encoding and decoding. These features are extracted once and stored on non-volatile memory elements, allowing for in-memory computation on neuromorphic platforms that prioritize specialized operations and energy-efficient processing. The Locally Competitive Algorithm (LCA)~\cite{rozell2008sparse} is used as the encoder to ensure that only a limited number of neurons are active at any given time, enabling sparsity and efficient coding of data. We evaluated PointLCA-Net on three datasets: NMNIST~\cite{orchard}, DVS128~\cite{amir2017low}, and Spiking Heidelberg Digits (SHD)~\cite{cramer2020heidelberg}, demonstrating how it can be uniformly applied to each of these datasets in a way that is deployable on neuromorphic platforms. We achieved high recognition accuracy across all these datasets while maintaining low computational overhead and energy efficiency, making the approach well-suited for deployment on neuromorphic systems.

The contributions of this work include the following:
\begin{itemize}
    \item A neuromorphic and energy-efficient adaptation of the PointNet and PointNet++ models.
    \item First application of the event cloud approach to neuromorphic datasets NMNIST and SHD.
    \item A proof of concept for an algorithm that can be uniformly applied to various spatio-temporal datasets for energy-efficient signal recognition on neuromorphic hardware.
\end{itemize}

\section{Related Work}
\label{sec:realted}

Only a modest number of recent works have studied how to apply SNNs to point clouds, primarily due to the prohibitive computational overhead involved in training. 

Spiking PointNet~\cite{ren2024spiking} is the first spiking version of PointNet, which leverages spiking neurons. The model is trained using a single time step, with membrane potential perturbation introduced to improve its performance. It provides some speedup and storage savings during the training phase and, in certain experiments, surpasses its ANN counterpart, suggesting the potential of SNNs for point cloud tasks. However, training with a single time step limits the model's ability to capture complex temporal dynamics, restricting its generalizability to new datasets. Additionally, the inherent complexity of training SNNs in this manner may become prohibitive when applied to more complex datasets beyond the ModelNet10 and ModelNet40 datasets, which are the only ones tested in the Spiking PointNet work. Unlike the structured 3D object data in the ModelNet datasets, DVS128 consists of unstructured spatio-temporal event data, which presents a different set of challenges. This method has not been evaluated on the DVS128 or other datasets considered in our work.

SpikePoint~\cite{renspikepoint} uses the PointNet++ architecture and converts point coordinates into spikes via rate coding. It employs surrogate backpropagation for training the network and has primarily been tested on event camera action recognition datasets. The use of surrogate gradients in this approach involves computing the derivative of a surrogate function, which can be computationally expensive, particularly for complex architectures such as PointNet++. The accuracy and convergence of training with surrogate gradients is also sensitive to the choice of surrogate function's parameters, potentially leading to additional computational costs for tuning and optimization.

\begin{figure*}[ht]
\centering
\includegraphics[width=1\textwidth]{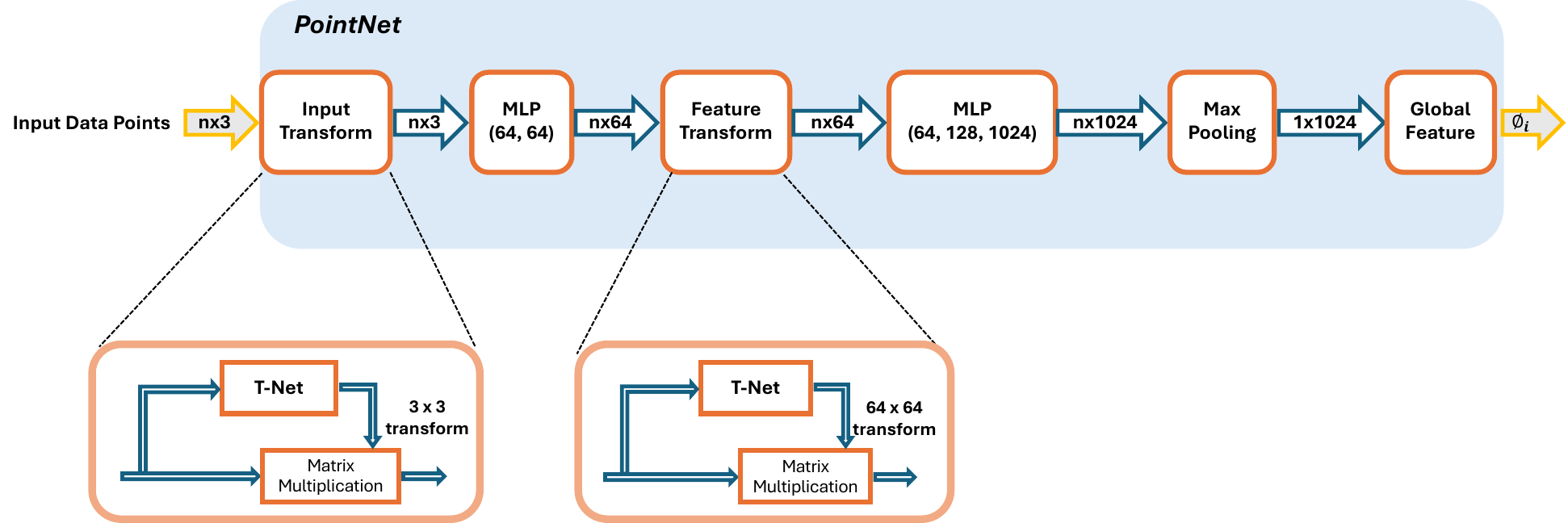}
\caption{Feature Extraction using PointNets.}
\label{fig:pointnet}
\end{figure*}

\section{PointLCA-Net}
Fig.~\ref{lca-arch} illustrates the architecture of PointLCA-Net. The training dataset is first converted into point clouds consisting of (x, y, t) point sets, pre-processed, and then fed into PointNet for feature extraction. These features are used to construct a dictionary, which is utilized by the Exemplar LCA-Decoder~\cite{mahmoodi2024d} to encode and then decode unseen test point clouds.

\subsection{Feature Extraction}
Fig.~\ref{fig:pointnet} illustrates the feature extraction process using a pre-trained PointNet~\cite{qi2017pointnet}, which processes point clouds by first aligning the input data points into a canonical space using an input transformation network (Input Transform). Each data point is then independently processed through a shared Multi-Layer Perceptron MLP(64, 64), feature transformation network (Feature Transform) and MLP(64, 128, 1024) to extract point-wise features. These features are aggregated using a symmetric Max Pooling function, to produce a permutation-invariant global feature vector $\phi_i$. The resulting 1024-dimensional feature vector captures comprehensive geometric and semantic information from the point cloud. These feature vectors are then used to construct the dictionary $\phi$ (see Fig.~\ref{lca-arch}), which is later used by the Exemplar LCA-Decoder algorithm (see Eq.~\ref{dictionary})).

\subsection{Overview of the Exemplar LCA-Decoder}
\label{sec:design}

The Exemplar LCA-Decoder algorithm~\cite{mahmoodi2024d} combines the sparse coding algorithm~\cite{rozell2008sparse} with the LCA algorithm~\cite{4379981} to encode the input signal $\mathit{S}$ as the activations $a_{i}$ of the Leaky Integrate and Fire (LIF) neurons where $\phi$ denotes a dictionary of features ($\phi_i$). The term $\varepsilon$ accounts for Gaussian noise in the reconstruction:

\begin{equation}
\label{recon}
S = \sum_{i=0}^{M-1} \phi_{i} a_{i}  + \varepsilon
\end{equation}

The membrane potential of the LIF neuron at time step $k$, denoted as $u_i[k]$, is influenced by a driving excitatory input $b_i$ and an inhibition matrix (Gramian) $G$. The Gramian matrix enables stronger neurons to suppress weaker ones, thereby promoting a sparse coding.

\begin{equation}
\label{neuron update}
\tau \dot{u}_i[k]+u_i[k]=b_i-\sum_{m\neq i}^{M-1}G_{i,m}a_m[k]
\end{equation}

\begin{equation}
\label{b_i}
b_i = S\phi_i
\end{equation}

\begin{equation}
\label{grammian}
G = \phi^{T}\phi
\end{equation}

\begin{equation}
\label{dictionary}
\phi = [\phi_0, \phi_1, \ldots, \phi_{M-1}]
\end{equation}

In PointLCA-Net, each $\phi_i$ represents a feature learned from a training point cloud. The thresholding function is:

\begin{align}
\label{thresholding}
    a_i[k] &= T_\lambda(u_i[k]) \nonumber \\
    &=
    \begin{cases}
        u_i[k] - \lambda \operatorname{sign}(u_i[k]), & \quad \left| u_i[k] \right| \geq \lambda \\
        0, & \quad \left| u_i[k] \right| < \lambda
    \end{cases}
\end{align}

Here, the threshold $\lambda$ represents the level the membrane potential must exceed for the neuron to spike. Next, a decoder is designed to decode the sparse codes $a_i$ and map them to \textit{K} distinct classes from the training dataset. This same decoder is then used to generate classifications for unseen test point clouds.

The Maximum Activation Code (Eq.\ref{max-decoder}) and Maximum Sum of Activation Codes (Eq.\ref{max-sum-decoder}) were used as decoders for classification. In contrast to a shallow (single layer) neural network decoder, which requires additional training and error backpropagation, the Maximum Activation Code and Maximum Sum of Activation Codes offer computational advantages and greater efficiency.

\begin{equation}
\label{max-decoder}
C_{Test} = Class[\underset{i}{argmax} [a_{i}]],
\end{equation}

\begin{equation}
\label{max-sum-decoder}
C_{Test} = Class[\underset{c}{argmax} (\sum_i \left |  a_{i}^{(c)}\right |)] 
\end{equation}

\vspace{10pt}

\section{Experiment Setup}
\label{sec:evaluation}
We evaluated PointLCA-Net using PyTorch~\cite{pytorch} on the following datasets: NMNIST, DVS128 and Spiking Heidelberg Digits (SHD). Events in these discrete spatio-temporal datasets were converted into point clouds, with each event represented by its spatial coordinates (x, y) and temporal component (t). We then concatenate these coordinates into a tensor of shape (x, y, t) for processing.  

\begin{figure}
\centering
\includegraphics[width=0.4\textwidth]{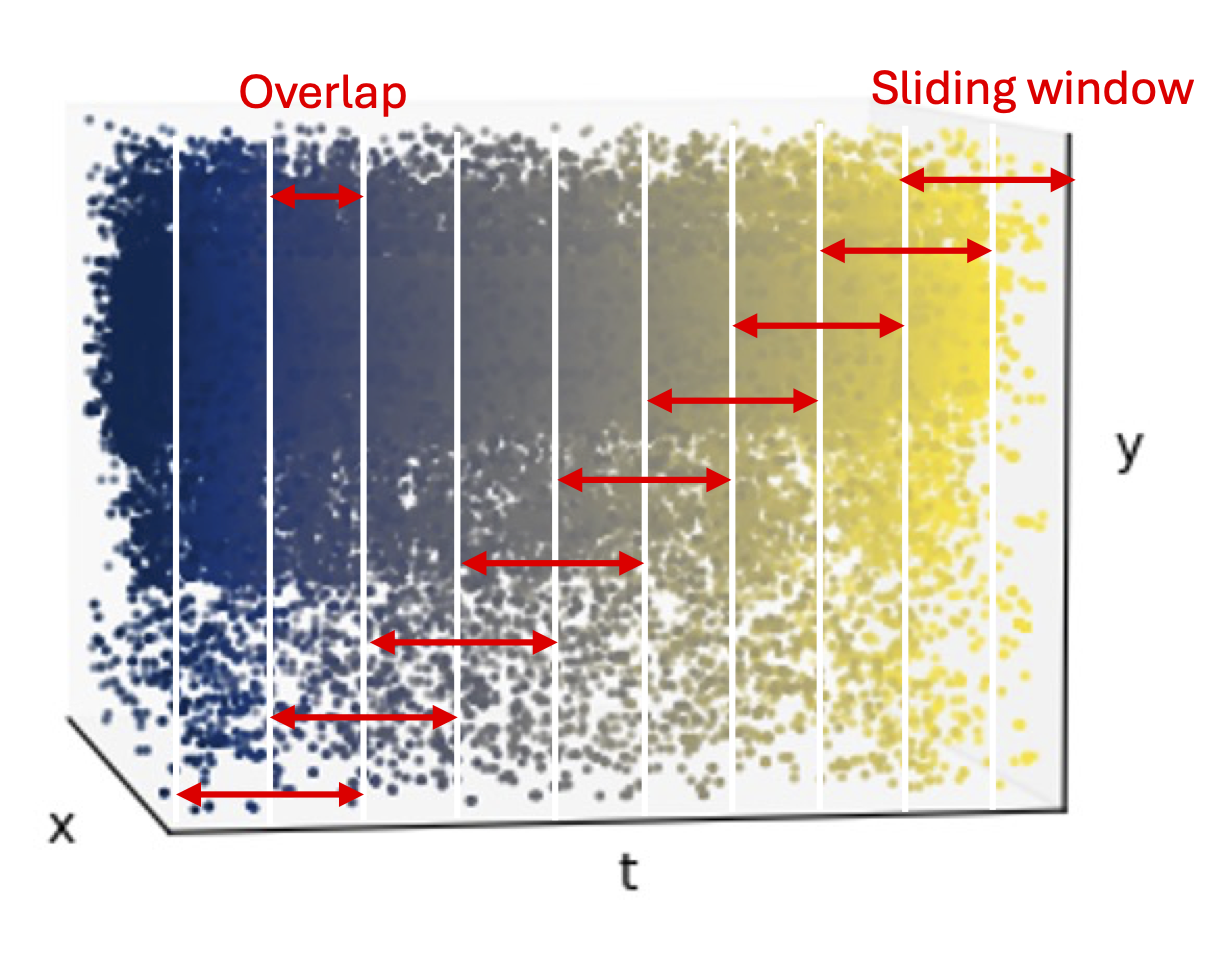}
\caption{Data Pre-processing}
\label{preprocessing}
\end{figure}

\subsection{NMNIST}
The NMNIST~\cite{orchard} dataset is a spiking variant of the MNIST dataset, generated by presenting images to a neuromorphic vision camera equipped with the ATIS (Asynchronous Time-based Image Sensor). This dataset comprises 60,000 training images, each consisting of 300 time samples, along with 10,000 test images used for accuracy evaluation.

\subsection{Spiking Heidelberg Digits (SHD)}
The SHD~\cite{cramer2020heidelberg} dataset comprises spoken digits from zero to nine in both English and German. Audio waveforms within the dataset are transformed into spike trains using an artificial cochlea model. The dataset includes 8,156 training samples and 2,264 test samples, each with a varying temporal duration.

\subsection{DVS128}
The DVS128~\cite{amir2017low} dataset, captured using a Dynamic Vision Sensor (DVS) camera, includes recordings of 11 distinct hand gestures, with approximately 20,000 to 40,000 events per gesture class. It contains 1,176 samples allocated for training and 288 samples for testing. These samples are further pre-processed, resulting in 28,606 training samples and 7,408 test samples. In our experiments, we included all 11 classes, including the random gesture class.

\vspace{10pt}

\subsection{Dataset Pre-Processing}
The preprocessing phase begins by evaluating the duration of the recordings. To account for the temporal variability in the dataset, we segment the data into sliding windows, with both the sizes and number of these windows determined based on the total duration of the recordings (see Fig.~\ref{preprocessing}). The data within each window is then shuffled and down-sampled to select a fixed number of events (128, 256, 512, or 1024). These sampled events are further concatenated, shuffled, and down-sampled, and normalized to create a final subset of 1024 events per data, which is subsequently input into the PointNets for feature extraction. 

\vspace{10pt}

\subsection{Workload and Energy Efficiency Analysis}
\label{Workload}
This section presents a thorough evaluation of the workload and energy efficiency of PointLCA-Net. We begin by estimating the number of Floating-Point Operations (FLOPs) to assess workload efficiency, followed by an analysis of its energy consumption. 
These estimations exclude the FLOPs required for feature extraction by PointNet and PointNet++. 


Let $N$ represent the global feature size, $M$ the dictionary size (which also corresponds to the number of neurons), and $K$ the total number of time steps. The number of active neurons, denoted as $\hat{M}$, represents the average number of firing neurons for which $a_i \neq 0$. The training and inference FLOPs are then calculated using the approach outlined in \cite{mahmoodi2024d} and \cite{takaghaj2024vit}:

\begin{align}
\label{FLOPS-training}
\underset{(Training)}{FLOPs} = \frac{M(M+1)(2N-1)}{2}
\end{align}

\begin{align}
\label{FLOPS2}
\underset{(Inferenec)}{FLOPs} = K(\frac{(2N-1)M}{K}+2M\hat{M}+M)
\end{align}

\begin{table}
\caption{Workload analysis of PointLCA-Net}
\label{flops}
\centering
\begin{tabular}{|l|>{\centering\arraybackslash}m{1.1cm}|l|l|l|l|>{\centering\arraybackslash}m{1.1cm}|}
\hline
 \textbf{Dataset} & \textbf{Training (TFLOPs)} &\textbf{K} & \textbf{N} & \textbf{M} & $\bm{\hat{M}}$ & \textbf{Inference (GFLOPs)}\\ \hline
  \multirow{2}{*}{NMNIST} & \multirow{2}{*} {3.7} & 100 & 1024 & 60K & 240 & 3 \\ \cline{3-7}
                            & & 10 & 1024 & 60K & 240 & 0.4 \\ \hline
                            
 \multirow{2}{*}{DVS128} & \multirow{2}{*} {0.84} & 100 & 1024 & 28.6K & 114 & 0.7 \\ \cline{3-7}
                            & & 10 & 1024 & 28.6K & 114 & 0.12\\ \hline

 \multirow{2}{*}{SHD} & \multirow{2}{*} {0.07} & 100 & 1024 & 8.156K & 33 & 0.07 \\ \cline{3-7}
                            &  & 10 & 1024 & 8.156K & 33 & 0.02 \\ 
\hline    

\end{tabular}
\end{table}

\begin{figure}
\centering
\includegraphics[width=0.4\textwidth]{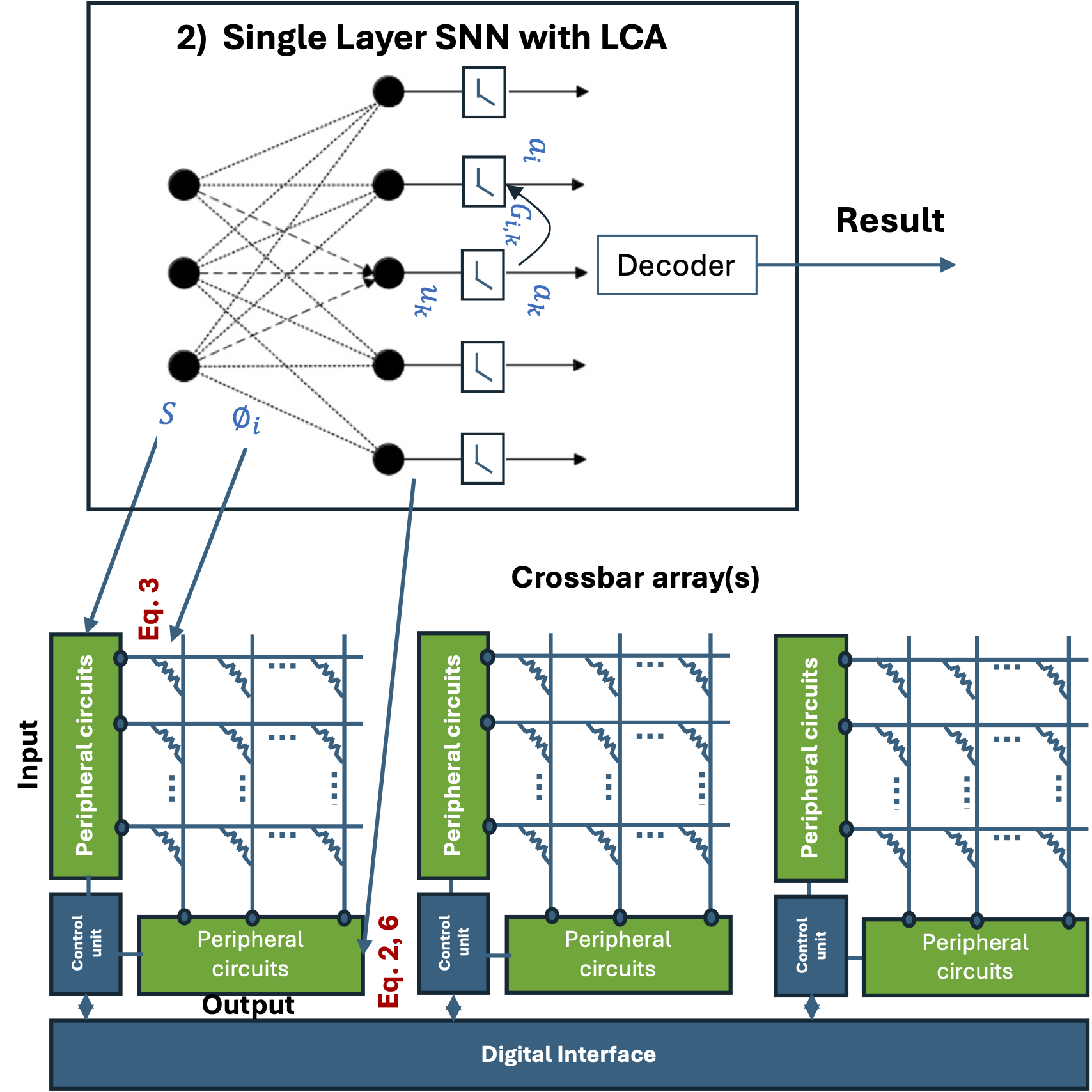}
\caption{PointLCA-Net Hardware Deployment}
\label{fig:p}
\end{figure}

\begin{table*}
  \centering
  \caption{Top-1 Test Accuracy Scores for PointLCA-Net.}
    \begin{tabular}{|c|c|c|c|c|c|c|}
    \cline{6-7}
        \multicolumn{3}{c}{} & \multicolumn{2}{c}{} & \multicolumn{2}{|c|}{\textbf{Accuracy}}  \\
    \cline{6-7}
    \multicolumn{3}{c}{} & \multicolumn{2}{c}{} & \multicolumn{2}{|c|}{Decoding Methods}  \\
      \hline
      \textbf {Dataset}& \textbf{Method} & \textbf{GFLOPs} & \textbf{Energy} & \textbf {Feature Extractor} & \textbf {$\underset{i}{max} \left\{ a_{i} \right\}$} & \textbf {$\underset{k}{max} \sum_{i} \left |  a_{i}^{(k)}\right |$} \\
    \hline
       NMNIST  & \textbf{PointLCA-Net} & 3 & 0.27 mJ &  PointNet & 95.49\% & 98.01\% \\
       &  &  &  &  PointNet++ & 94.74\% & \textbf{98.78\%} \\
      \hline   
     SHD& \textbf{PointLCA-Net} & 0.07 & $6.4 \, \mu$J &  PointNet & 59.85\% & 65.11\% \\
       &  &  &  &  PointNet++ & 66.30\% & \textbf{78.46\%} \\
     \hline      
    DVS128& \textbf{PointLCA-Net} & 0.7& 0.065 mJ &  PointNet & 82.67\% & 90.02\% \\
       &  &  &  &  PointNet++ & 86.69\% & \textbf{93.41\%} \\
      \hline

     \hline
       DVS128& PointNet~\cite{wang2019space}$^{\mathrm{a}}$ & - & - & - &  \multicolumn{2}{c|}{90.20\%$^{\mathrm{c}}$} \\ 
       & PointNet++~\cite{wang2019space}$^{\mathrm{a}}$ & 0.872$^{\mathrm{b}}$ &  4.01 mJ (Dynamic) and  & - &  \multicolumn{2}{c|}{94.10\%} \\ 
    & & &  1.936 mJ (Static)$^{\mathrm{b}}$ &  &  \multicolumn{2}{c|}{} \\
         \hline
       DVS128& SpikePoint~\cite{renspikepoint}$^{\mathrm{a}}$ & 0.9& 0.82 mJ (Dynamic) and &  - &  \multicolumn{2}{c|}{98.1\%$^{\mathrm{d}}$} \\  
        & & & 0.756 mJ (Static) &  &  \multicolumn{2}{c|}{} \\  
\hline 
\multicolumn{7}{l}{$^{\mathrm{a}}$These methods haven't been tested on NMNIST or SHD.}\\
\multicolumn{7}{l}{$^{\mathrm{b}}$ The values are taken from the results reported in~\cite{renspikepoint}.}\\
\multicolumn{7}{l}{$^{\mathrm{c}}$ This accuracy was achieved using 10 classes, excluding the random gestures class.}\\
\multicolumn{7}{l}{$^{\mathrm{d}}$This accuracy was achieved using a different data partitioning and sample size than ours.}\\
\end{tabular}
\label{tab:decoders}
\end{table*}

\begin{table}
\caption{Hyperparameters}
\label{parameter}
\begin{center}\small
\begin{tabular}{|c|c|c|}
\hline
\textbf{Symbols} & \textbf{Description} & \textbf{Value} \\
\hline
$\lambda$ & Threshold & 0.2 \\
$\tau$ & Leakage &  1000\\
$k$ & Number of time steps &  100\\
\hline
\hline
\multirow{3}{*}{\textbf{Data Pre-Processing}} & Sliding window & 0.5 sec\\
                                       & Overlap & 0.25 sec\\
                                       & Number of event points & 1024 \\
\hline
\end{tabular}
\end{center}
\end{table}

\vspace{10pt}
Table~\ref{flops} presents the estimated training and inference FLOPs for PointLCA-Net. ``TFLOPs" refers to the Tera FLOPs required for training, while ``GFLOPs" indicates the Giga FLOPs for inference operations. According to \cite{mahmoodi2024d}, up to 0.4\% of neurons spike at each time step, and we have verified that this estimate is accurate. The spiking sparsity introduced by LCA encoding resulted in an average reduction of 99.54\% in computational effort during inference. Furthermore, we explored two scenarios with 100 and 10 time steps (\textit{K}). Reducing the number of time steps from 100 to 10 led to an average decrease of 80\% in inference FLOPs.

\subsection{Neuromorphic Hardware Mapping}
\label{Hardware}
The neuromorphic-friendly architecture of the Exemplar LCA-Decoder supports the mapping of PointLCA-Net to low-energy neuromorphic platforms, enabling the use of energy-efficient in-memory computing. Using RRAM crossbar arrays, Yao et al.~\cite{yao2020fully} demonstrated an energy efficiency of 11 Tera FLOPs per watt for MAC operations, which corresponds to an energy consumption of approximately \( 9.09 \times 10^{-14} \) joules per floating-point operation.
Beyond improving workload efficiency through sparse coding, PointLCA-Net can also benefit from recent advancements in in-memory computing~\cite{sebastian2020memory} and memristive crossbar arrays to further enhance energy efficiency. 
Mahmoodi et al.~\cite{takaghaj2024vit} proposed a hardware implementation of the Exemplar LCA-Decoder algorithm in ViT-LCA. Similarly, Fig.\ref{fig:p} depicts the hardware mapping of PointLCA-Net. In this architecture, the values of $\phi_{i}$ are represented as the conductance of memristors in each column, which are used to compute the neuron excitatory inputs, as described in Eq.\ref{b_i}. Input data is preprocessed in the input peripheral circuits, while neuron dynamics and thresholding (defined in Eq.\ref{neuron update} and Eq.\ref{thresholding}) are handled in the output peripheral circuits of the neural processor.

\section{Evaluation}

Table~\ref{tab:decoders} presents the energy estimations and test accuracy for all three datasets tested. All experiments were conducted using the hyperparameters specified in Table~\ref{parameter}, selected to achieve an accuracy near 100\% when tested on the training data. The results in Table~\ref{tab:decoders} indicate that the ``Maximum Sum of Activations" decoder outperforms the ``Maximum Activation" decoder, delivering significantly better performance across all datasets. Additionally, using PointNet++ for feature extraction yielded better results than PointNet. The energy estimation was calculated based on the estimated GFLOPs during inference (Table~\ref{flops}) and the expected $9.09 \times 10^{-14}$ joules energy consumption per floating point operation when the operations are mapped onto the RRAM memory crossbar arrays. Wang et al.~\cite{wang2019space} reported an accuracy of 90.20\% on the DVS128 dataset with 10 classes when trained directly on PointNet, and 94.10\% when trained using PointNet++. Their work was the first to apply PointNet and PointNet++ to spatio-temporal data; however, they did not address hardware deployment nor use a spiking approach, leaving the energy efficiency of the approach unknown.

Furthermore, SpikePoint~\cite{renspikepoint}, a spiking variant of PointNet++, has been tested on the DVS128 dataset. SpikePoint uses a surrogate-based backpropagation method, achieving an accuracy of 98.6\% with a GFLOPs value of 0.9, compared to 0.7 for PointLCA-Net. According to the authors, their study used 26,796 training samples and 6,959 test samples, with the test set comprising ``20\% randomly selected from the total samples". In contrast, the DVS128 dataset~\cite{amir2017low} used in this study contains 28,606 training samples and 7,408 test samples, with the test data kept entirely separate from the training data. Given the differences in data partitioning and sample size, directly comparing accuracy may not be fully representative. Additionally, PointLCA-Net is significantly more energy-efficient, with a total energy consumption of 0.065 mJ—approximately 24.23 times lower than that of SpikePoint, which has a reported dynamic energy consumption of 0.82 mJ and a static energy consumption of 0.756 mJ. All energy calculations are based on theoretical estimations and calculations.

\section{Discussion}
\label{}

This study serves as a proof of concept and has significant potential as an algorithm that can be uniformly applied to various spatio-temporal data beyond DVS128, including video and audio. Our focus was on leveraging PointNet and PointNet++ in PointLCA-Net to recognize spatio-temporal signals in an energy-efficient manner that could be mapped onto current neuromorphic hardware, rather than solely enhancing their accuracy. PointLCA-Net currently has limitations in accuracy compared to state-of-the-art methods, and further improvements could be made by exploring alternative architectures for feature extraction, such as Point Transformer~\cite{zhao2021point}. 

PointLCA-Net does not require any additional training of the PointNet and PointNet++ models used for feature extraction. Once the features are extracted from the training data, a dictionary is constructed and used in the Exemplar LCA-Decoder.

Finally, in this study, we utilized the entire set of training data points to construct the dictionary. While larger dictionary lengths can improve accuracy, they may also create a computational burden for larger datasets. To enhance computational efficiency, one potential approach is to reduce the dictionary length \( M \) through Principal Component Analysis (PCA), Support Vector Machines (SVM), or by selecting discriminative features, thereby decreasing the computational overhead. For instance, a 10\% reduction in dictionary length could, on average, reduce inference FLOPs by 17\% and training FLOPs by 20\%. As a result, this opens up new avenues for enhancing computational efficiency and model performance of PointLCA-Net in future studies.

\section{Conclusion}
\label{sec:conclusion}

In this work, we explored the integration of Exemplar LCA with PointNets for the first time and evaluated its performance on classification tasks using the NMNIST, DVS128, and SHD datasets.
We utilized features extracted through PointNet and PointNet++ within an Exemplar LCA encoder-decoder framework, which eliminated the need for training a dictionary as done in the original LCA. Additionally, we demonstrated how this approach can be mapped to memristor crossbar arrays for efficient in-memory computing on edge devices. This study highlights the potential for further research, including the investigation of alternative architectures such as Point Transformer for even higher performance. Given the energy efficiency potential of PointLCA-Net and its adaptability to various datasets, it presents an opportunity for enhancement in spatio-temporal signal recognition.

\section{Acknowledgement}
The author would like to thank Dr. Jack Sampson for his valuable guidance and constructive feedback during the preparation of this manuscript and Harm Munk for his assistance with the image used in Fig. 2.

\vspace{12pt}

\end{document}